%% file: main.tex
\documentclass{llncs}
\usepackage{booktabs} 

\usepackage{array}
\newcolumntype{L}{>{\centering\arraybackslash}m{3.0cm}}
\newcolumntype{C}{>{\centering\arraybackslash}m{2.0cm}}
\usepackage{graphicx}
\usepackage{cite}
\usepackage[misc]{ifsym}

\begin{document}
	\title{''Is this an example image?'' -- Predicting the Relative Abstractness Level of Image and Text} 
	\author{Christian Otto\textsuperscript{(\Letter)}\inst{1,2}\orcidID{0000-0003-0226-3608}\and Sebastian Holzki\inst{2}\orcidID{0000-0002-1636-711X}\and Ralph Ewerth\inst{1,2}\orcidID{0000-0003-0918-6297}}
	\institute{Leibniz Information Centre for Science and Technology (TIB), Welfengarten 1B, 30167 Hannover, Germany\newline\email{first.lastname@tib.eu} \and L3S Research Center, Leibniz Universit\"at Hannover, Germany}

\maketitle

\begin{abstract} te the feasibility of the approach. 

Successful multimodal search and retrieval requires the automatic understanding of semantic cross-modal relations, which, however, is still an open research problem. Previous work has suggested the metrics \textit{cross-modal mutual information} and \textit{semantic correlation} to model and predict cross-modal semantic relations of image and text. In this paper, we present an approach to predict the (cross-modal) relative abstractness level of a given image-text pair, that is whether the image is an abstraction of the text or vice versa. For this purpose, we introduce a new metric that captures this specific relationship between image and text at the \textit{Abstractness Level (ABS)}. We present a deep learning approach to predict this metric, which relies on an autoencoder architecture that allows us to significantly reduce the required amount of labeled training data. For this, a comprehensive set of publicly available scientific documents has been accumulated. Experimental results on a challenging test set demonstrate the feasibility of the approach. 

\end{abstract}

\input{1_introduction.tex}
\input{2_related_work.tex}

\input{3_metrics.tex}
\input{4_our_approach.tex}
\input{5_experiments.tex}
\input{6_conclusion.tex}

\bibliographystyle{splncs04}
\bibliography{references}

\end{document}

%% file: 1_introduction.tex
\section{Introduction}
\label{sec:introduction}

In this era of big data, the proliferation of multimodal web content in online news, social networks, open educational resources, video portals, etc. is increasing drastically. Graphics and pictures in multimodal documents are a powerful communication channel to illustrate, decorate, detail, summarize, or complement textual information. This is particularly true for educational and scientific material. In this context, graphical and pictorial information can be very important to support learning scenarios as, for instance, in the recently evolved field of search as learning. To enable truly multimodal recommender systems for web search, an automatic understanding of the \textit{multimodal} content and the inherent cross-modal relations are a prerequisite. In this respect, however, information retrieval research did not address all possible kinds of cross-modal relations between images and text in a differentiated way\footnote{In contrast to research in communication sciences and applied liguistics where the visual/verbal divide has been researched in a very detailed way for decades.}. Typically, (multimedia) information retrieval research assumes a semantic correlation \textit{in general} in case image and text are placed jointly on purpose. An exception is Henning and Ewerth's work~\cite{henning2017estimating} that suggest two metrics (dimensions) to differentiate image-text relations by 1.) cross-modal mutual information (CMI) and 2.) semantic correlation (SC).   

In this paper, we argue that these two metrics do not completely cover all possible types of image-text relations, particularly when considering educational or scientific content. Therefore, we suggest an additional metric: the relative \textit{Abstractness Level} (ABS) that measures whether an image depicts information of a related text at a more detailed or a more abstract level, or at the same level. Furthermore, we propose a deep learning approach to automatically predict the abstractness level of a given image-text pair. The system relies on an autoencoder architecture and multimodal embeddings. Since the deep learning system requires a sufficiently large amount of training data, we have gathered an appropriate dataset from a variety of Web resources. Experimental results on a demanding test set demonstrate the feasibility of the proposed system.

The paper is structured as follows. Related work is summarized in Section 2 from the perspectives of communication sciences and information retrieval. 
Section 3 motivates the new metric of abstractness level and explains the proposed deep learning system to automatically predict this metric, while Section 4 describes the data acquisition process for the training of the deep networks. The experimental results are presented in Section 5. Section 6 concludes the paper and outlines areas of future work.

%% file: 2_related_work.tex
\section{Related Work}
\label{sec:relatedwork}
\subsection{Image-Text Relations and the Visual/Verbal Divide}
\begin{figure}[htbp]
	\centering
	\includegraphics[width=0.8\textwidth]{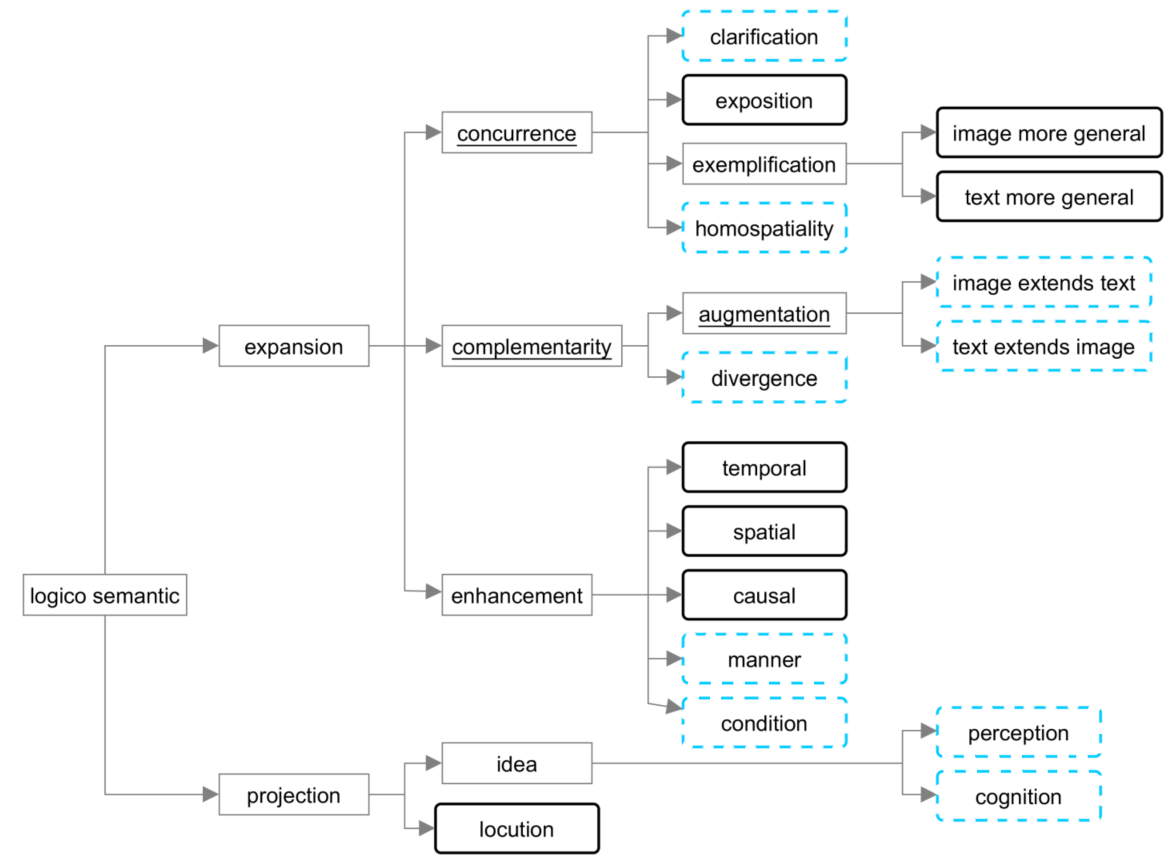}
	\caption{The logico-semantics part of Unsworth'S taxonomy~\cite{unsworth2007image} is shown, where blue borders show extensions to Martinec and Salway~\cite{Martinec2005} and underlined names were changed by the authors, but have the same meaning.}
	\label{fig:unsworth}
\end{figure}

The interplay between visual and textual\footnote{Textual information can be considered as visual information as well, of course. Here, we denote graphical and pictorial information as \textit{visual information}.} information has been subject to research for decades in the fields of communication sciences and applied linguistics. One of the early attempts to comprehensively categorize the joint placement of images and text date back to Barthes~\cite{barthes1977image}, who set the groundwork for a lot of categorizations that developed later. For example, Martinec and Salway~\cite{Martinec2005}, Marsh and White~\cite{marsh2003taxonomy}, and Unsworth~\cite{unsworth2007image} build upon Barthes' taxonomy, which defines the \textit{Status} relation between an image and its accompanied text. It describes if there is a hierarchical dependency between both modalities or if they are equally important in conveying the information intended by the author. The aforementioned taxonomies extend this distinction with different interpretations of Halliday's~\cite{halliday2013halliday} \textit{logico-semantics}, which are a linguistic method to describe different types of text clauses. The application of these fine-grained distinctions of the logico-semantics to image-text pairs result in very detailed taxonomies. Figure \ref{fig:unsworth} shows the latest version of Unsworth's extensions to Martinec and Salway's taxonomy \cite{Martinec2005}. 
While these taxonomies are comprehensive, their level of detail makes it sometimes difficult to assign an image-text pair to a particular class, as criticized by Bateman~\cite{bateman2014text}, for instance. Recent work from the field of multimedia retrieval approaches this problem differently through two metrics that are more general and easier to infer: Henning and Ewerth~\cite{henning2017estimating}, 
suggest the following two metrics (or dimensions):
1.) \textit{Cross-modal Mutual Information (CMI)} is defined as the amount of shared entities or concepts in both modalities, ranging from 0 to 1;
and 2.) \textit{Semantic Correlation (SC)} is defined as the amount of shared meaning or context, indicating if the information contained in both modalities are aligned, uncorrelated or contradictory, i.e., ranging from -1 to 1. 
They show that a deep learning approach that utilizes multimodal embeddings can basically predict these interrelation metrics. 

\subsection{Machine Learning for Multimodal Data Retrieval}
In this subsection, a brief overview of methods to encode heterogeneous modalities for machine learning and multimedia retrieval approaches is given. There are several possibilities to encode data samples consisting of distinct modalities~\cite{baltruvsaitis2018multimodal}. The choice of the optimal method depends on multiple factors: the type of modality to encode, the number of training samples available, the type of classification to perform and the desired interpretability of the models. One type of algorithms utilizes \textit{Multiple Kernel Learning}, which is an extension of kernel-based support vector machines~\cite{bucak2014multiple, gonen2011multiple}. They consist of a kernel specifically designed for each modality and thus allow for the fusion of heterogeneous data. Application domains are, for instance, multimodal affect recognition~\cite{poria2015deep, jaques2015multi}, event detection~\cite{yeh2012novel}, and Alzheimer's disease classification~\cite{liu2014multiple}. An advantage is their flexibility in the kernel design and global optimum solutions, but they have a rather slow inference time.
Deep neural networks are another technique to model multiple modalities at once. Due to their growing popularity in recent years, there is also much research on designing deep learning systems for processing multimodal data. Such research directions include approaches for audio-visual~\cite{afouras2018deep, meutzner2017improving}, audio-gesture~\cite{neverova2016moddrop} and textual-visual~\cite{jin2016video, Ramanishka2016} data. The common idea is to encode each modality individually and fuse them in joint hidden layers. Especially well suited are these methods for encoding temporal information like sentences, which fits the problem addressed in this paper nicely. For example, Cho et  al.~\cite{cho2014learning} use Gated Recurrent Units (GRU) to encode sentences, but it is also possible to utilize Long-Short-Term Memory (LSTM) cells~\cite{vinyals2015show}.
More recent extensions are shown by Jia et al.'s~\cite{jia2015guiding} and Rajagopalan et al.'s~\cite{rajagopalan2016extending} works, which model temporal information for textual as well as for the visual components. Neural networks are also able to learn meaningful embeddings of multimodal data in an unsupervised manner via an autoencoder architecture, which not only removes the necessity for hand-crafted features but also can significantly reduce the required amount of labeled training data~\cite{henning2017estimating}.
Cross-media and multimedia retrieval is an area of research that profits the most from techniques to bridge the semantic gap between image and text~\cite{Shutova2016, balaneshin2018deep,kang2015cross,yan2016supervised}. Fan et al.~\cite{Fan2017} implement a multi-sensory fusion network, which improves the comparability of heterogeneous media features and is therefore well suited for image-to-text and text-to-image retrieval. A self-paced cross-modal subspace matching method constructing a multimodal graph is proposed by Liang et al.~\cite{Liang2016}. It is designed to preserve the intra-modality and inter-modality similarity between the input samples. Carvalho et al.~\cite{carvalho2018cross} proposed the \textit{AdaMine} model, which combines instanced-based and semantic-based losses for a joint retrieval and semantic latent space learning method. This method is utilized to retrieve recipes from pictures of food and vice versa.

%% file: 3_metrics.tex
\section{The Abstractness Metric for Image-Text Relations}
\label{sec:deducing}
In this section, we motivate and derive the new metric of relative abstractness for image-text pairs. In this respect, we analyze the existing gap in applied linguistics and communicaton sciences, as well as information retrieval. 

\subsection{Analysis}
\label{sec:observations}
\begin{figure}[htbp]
	\centering
	\includegraphics[width=0.8\textwidth]{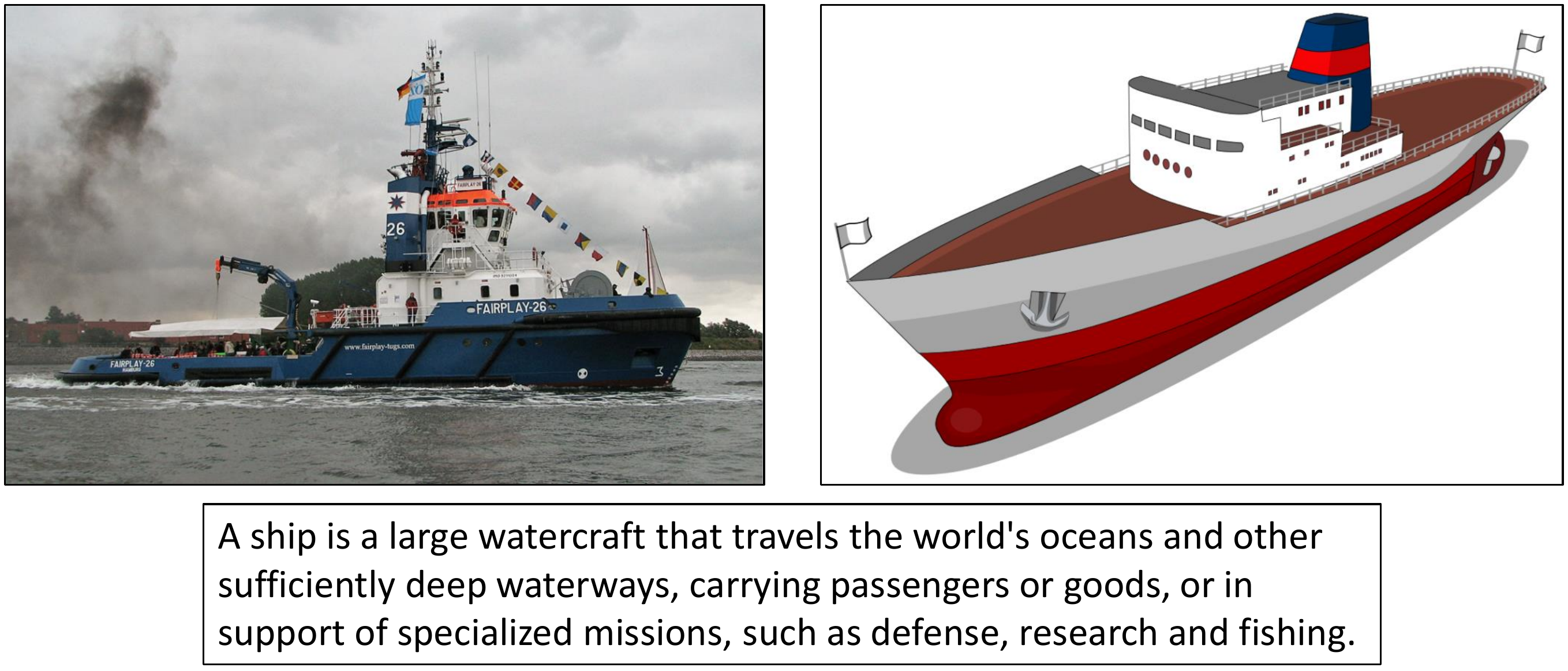}
	\caption{Image-text classes \textit{sample} and \textit{exemplify} by Marsh and White, showing that the authors use the concept of \textit{Abstractness Level} to add more depth to their categorizations.}
	\label{fig:ships}
\end{figure}

Marsh and White~\cite{marsh2003taxonomy} introduce three separate taxonomies that distinguish different levels (low, medium, high) of semantic relations for image-text pairs, which is similar to Henning and Ewerth's \textit{Semantic Correlation}. In other words, they distinguish between three different levels of the same metric, implying that the underlying sub-hierarchies themselves have a certain degree of overlap. Thus, it can be deduced that there is either an ambiguity issue or a missing property to make an actual distinction. With respect to some of the overlapping classes it can be observed that their distinction is based on the level of abstractness. 
An example is shown in Figure \ref{fig:ships} that portrays the classes \textit{sample} and \textit{exemplify} by Marsh and White~\cite{marsh2003taxonomy}.

It shows a textual phrase and two visual representations which together in both cases create an \textit{Illustration} example according to Barthes~\cite{barthes1977image}. However, they belong to Marsh and White's \textit{sample} and \textit{exemplify} classes, where the latter one is defined as an ideal example and the first one can be any concrete instance of the described concept. Therefore, the actual distinction is made by means of their abstractness, which is also a very important concept for scientific or educational material to improve comprehensibility. %
\subsection{Implications and the Abstractness Metric}
\label{sec:generality}
We claim that the \textit{relative difference of the Abstractness Level (ABS)} is an essential part in describing the relations between an image and text. To support this assumption, we list in Table~\ref{tab:generality} a number of image-text classes that contain a certain difference in abstractness by definition.
\begin{table}
	\begin{tabular}{| C | L | L | L |}
		\hline
		ABS $\rightarrow$ & \textbf{I $=_a$ T}& \textbf{I $>_a$ T} & \textbf{I $<_a$ T}\\
		Reference $\downarrow$ & & & \\
		\hline \hline
		Martinec \& Salway~\cite{Martinec2005} & exposition, locution, idea & image more general, enhancement by text & text more general, enhancement by image\\ 
		\hline
		Unsworth~\cite{unsworth2007image}  & exposition, clarification, locution, perception, cognition & text instantiates image, enhancement by text & image instantiates text, enhancement by image\\ 
		\hline
		Marsh \& White~\cite{marsh2003taxonomy}  & compare, contrast, concentrate, compact, model & sample, exemplify, isolate, contain, locate, induce perspective, emphasize, document & graph, translate, describe, define\\
		\hline
	\end{tabular}
	\centering
	\vspace{0.1cm}
	\caption{Overview of image-text classes that entail a certain difference on the abstractness level between image and text. (Note: $>_a$ is read as "is more abstract than")}
	\label{tab:generality}
	\vspace{-0.3cm}
\end{table}
This implies that a metric describing the relative difference in the abstractness level is indeed necessary to characterize an image-text relation. We would like to emphasize the term \textit{relative}, since it is important that image and text are considered jointly. Therefore, a particular image can be less abstract than a text, while it is more abstract than another one. Also, in order to differentiate between abstractness levels it is necessary to have an object of reference, or a \textit{Cross-modal Mutual Information} $CMI > 0$, as it is the case for the ship example in Fig. \ref{fig:ships}.

%% file: 4_our_approach.tex
\section{Predicting the Abstractness Level of Image and Text}
\label{sec:methodology}
In this section, we present a system that automatically measures the relative abstractness level between an image and its associated text. 
There are no repositories and Web resources of image-text pairs that can be easily exploited to train a deep learning classifier. Consequently, we follow an autoencoder approach similar to \cite{henning2017estimating}, which requires much less labeled data. We gather the necessary samples from open access publications provided by Sohmen et al.~\cite{sohmen2018figures}, as explained in detail in the next subsection. 

\subsection{Data Acquisition}
\label{sec:data}
For the purpose of legal re-use, Sohmen et al.~\cite{sohmen2018figures} provide illustrations of publications of the open access publisher Hindawi\footnote{https://www.hindawi.com/}. The majority of Hindawi articles is available under the Creative Commons Attribution License, so that they can be used for this type of research. Another advantage is that they are accessible in XML format, which makes them easier to read than files in PDF format. We crawled 288,057 image-text pairs from four different journals, namely \textit{Advances in Artificial Intelligence (AAI)}, \textit{Applied Computational Intelligence and Soft Computing (ACISC)}, \textit{Advances in Multimedia (AM)} and \textit{Mathematical Problems in Engineering (MPE)}. The final distribution of articles is presented in Table~\ref{tab:ae_input}.

\begin{table}[htbp]
	\begin{tabular}{| l | c  c  c |}
		\hline
	    Journal & \#articles & \#figures & \#image-text pairs \\ 
		\hline \hline
		AAI     & 94  & 1,217 & 3,180 \\
		ACISC   & 185 & 2,215 & 5,453 \\ 
		AM      & 144 & 2,304 & 6,057 \\
		MPE     & 8,251 & 106,435 & 273,367 \\
		\hline
		Sum     & 8,674 & 112,171 & 288,057\\ \hline 
	\end{tabular}
	\centering
	\vspace{0.1cm}
	\caption{Overview of the training data used for the autoencoder.}
	\label{tab:ae_input}
	\vspace{-0.3cm}
\end{table}
We manually labeled more than 3,000 image-text pairs for training and testing. The data distribution of the labeled data is presented in Table \ref{tab:labeledset}.
\begin{table}[htbp]
	\begin{tabular}{| l | c c c | c c |}
		\hline
	    Journal & T $>_{a}$ I & T $<_{a}$ I & T $=_{a}$ I & Sum & Percentage \\ 
		\hline \hline
		AAI     & 322  & 113 & 145 & 580 & $19.2\%$ \\
		ACISC   & 383  & 173 & 242 & 798 & $26.5\%$ \\
		AM      & 354  & 169 & 169 & 858 & $28.4\%$ \\
		MPE     & 352  & 255 & 255 & 780 & $25.9\%$ \\
		\hline
		Sum     & 1,411 & 710 & 895 & 3,016 & - \\ 
		Percentage & $46.8\%$ & $23.5\%$ & $29.7\%$ & - & - \\ \hline 
	\end{tabular}
	\centering
	\vspace{0.1cm}
	\caption{Overview of the manually labeled part of the data which is used to train and evaluate the classifier network.}
	\label{tab:labeledset}
	\vspace{-0.3cm}
\end{table}

Our annotation process results in a minimum number of 700 samples per image-text class which is sufficient to train a classifier that uses the embeddings of our pre-trained autoencoder network.
\subsection{Representing Multimodal Data via Autoencoding}
\label{sec:autoencoder}
\begin{figure}[htbp]
	\centering
	\includegraphics[width=0.9\textwidth]{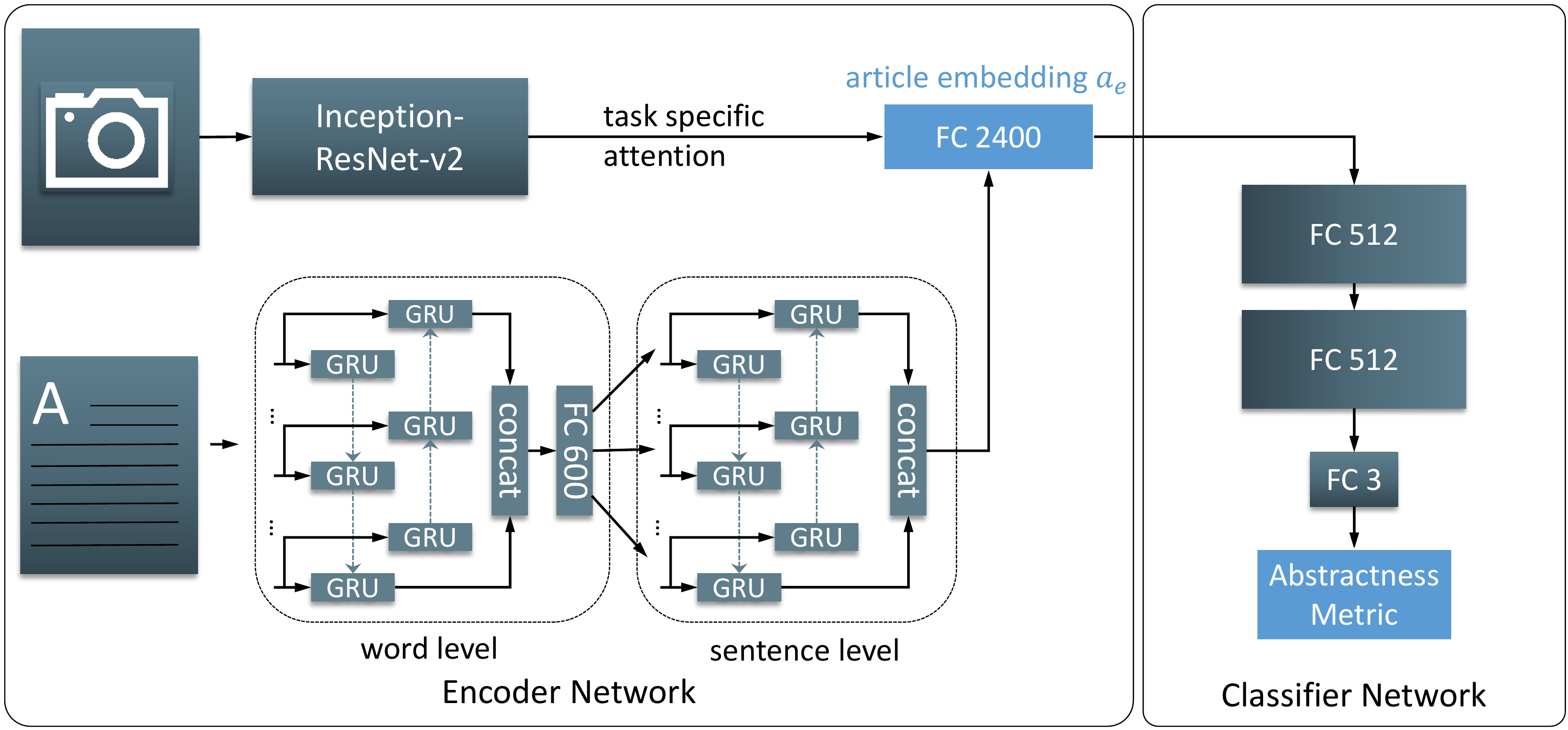}
	\caption{Overview of the encoder and classifier network.}
	\label{fig:encoder_classifier}
	\vspace{-0.3cm}
\end{figure}

We suggest an autoencoder approach for two main reasons: First, the automatic generation of labeled training data is not possible since the available amount of annotated image-text pairs is limited. It is not reasonable to train a classifier from scratch with less than 1,000 samples per class. Second, the encoder-decoder architecture allows for adjustments that fit nicely to our scenario and also allows us to investigate if the right information is preserved by the embedding.
Our design is similar to Henning and Ewerth's~\cite{henning2017estimating} approach, but includes some modifications that consider the nature of figures and illustrations in scientific documents as opposed to natural images. Also, we replace some components in the encoding part with more recent system components, see Figure \ref{fig:encoder_classifier}. In detail, we use a pre-trained model of the Inception-ResNet-v2~\cite{szegedy2017inception} without its classification layers to encode the input image as well as the preprocessing pipeline suggested by Szegedy et al.~\cite{szegedy2015going}. The textual information is preprocessed by removing any specific XML characters and by replacing formulas with the word "formula". In addition, we truncate sentences that are longer than 50 words and paragraphs longer than 30 sentences. The resulting feature vector of image encoding is then fed into the attention mechanism of the text encoding step, where, inspired by Yang et al.~\cite{yang2016hierarchical}, a bidirectional recurrent neural network (RNN) architecture is used consisting of multiple GRU cells. This way the text is encoded in a hierarchical way: first on a sentence and then on a full-text level. After concatenating the image and text embedding we receive a 2,400 dimensional \textit{article embedding} according to Henning and Ewerth \cite{henning2017estimating} (Figure \ref{fig:encoder_classifier}).

To obtain a high-quality multimodal embedding for image-text pairs, we aim at a decoder that reconstructs the image as well as text from the encoded article embedding. We compute a loss between input and output information that describes how well image and text can be reproduced from the condensed representation. A first fully-connected layer decides which parts of the embedding are important to reproduce the image and therefore generates a first 30x30 predicted reconstruction of the visual data. An alternating series of up-scaling and convolutional layers subsequently produces an image that corresponds to the size of the input image (300x300). In contrast to Henning and Ewerth~\cite{henning2017estimating}, the convolutional layers use a kernel size of $3x3$ instead of $5x5$, which is necessary to successfully reproduce the fine lines depicted in many scientific tables or diagrams. Another difference is the size of the convolutional layers in the pipeline, where we use (128, 64, 32, 3) opposed to (32, 8, 3). The loss between input and output image is computed using mean squared error.

\begin{figure}[htbp]
	\centering
	\includegraphics[width=0.8\textwidth]{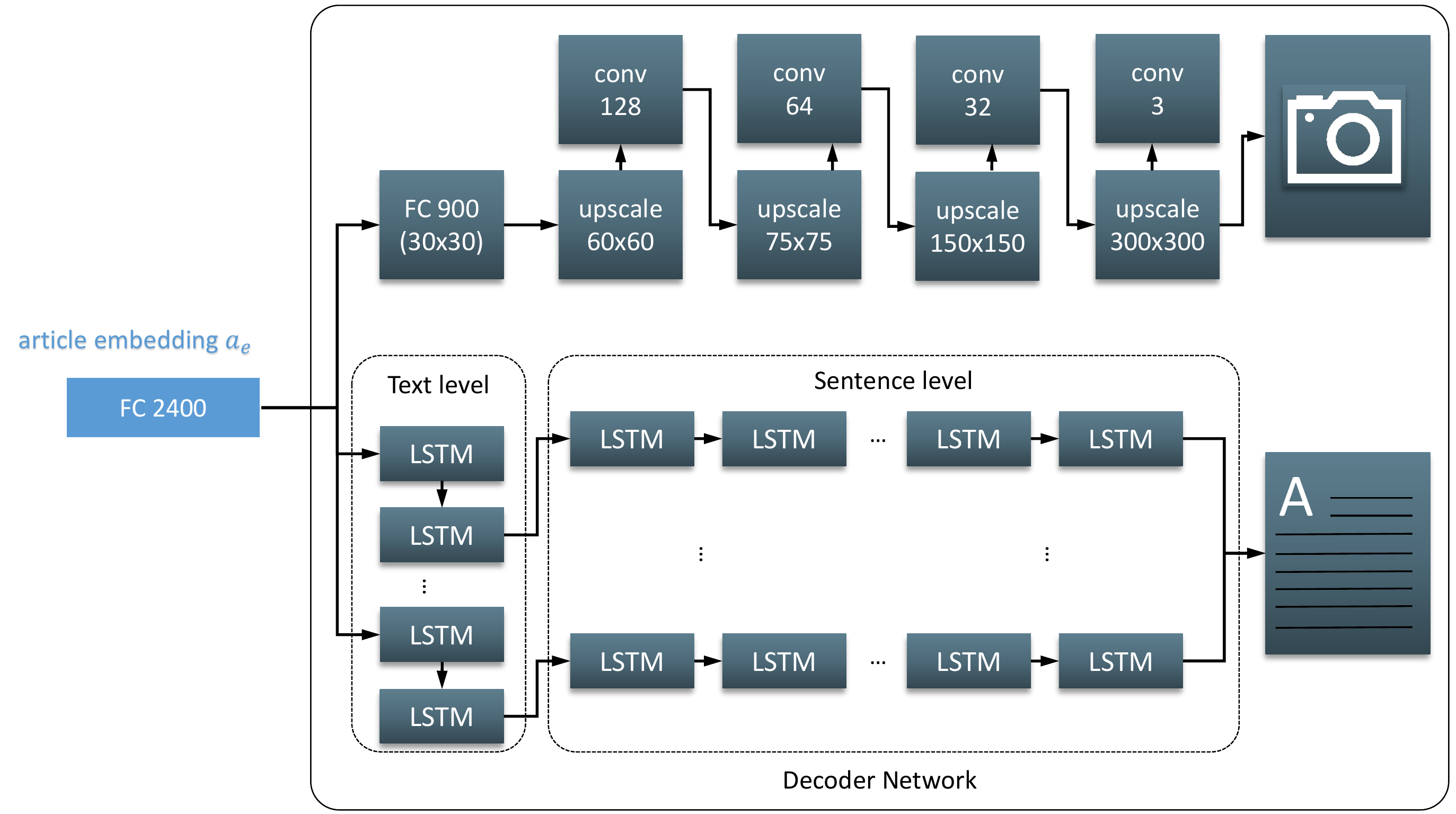}
	\caption{Overview of the decoder network, whose input is the article embedding generated by the encoder (Figure \ref{fig:encoder_classifier}).}
	\label{fig:decoder}
\end{figure}

The textual information is reconstructed by a hierarchical unidirectional RNN consisting of LSTM cells, which proved to be more powerful than Gated Recurrent Units (GRUs) for this task. It first generates sentence features from the article embedding and uses them afterwards on a word level to estimate the original input text. Both hierarchy layers use the batch normalization technique proposed by Ba et al.~\cite{ba2016layer}. An overview of the decoder network can be seen in Figure \ref{fig:decoder}. The loss between input and output is computed based on a word embedding that is based on a predefined fastText \cite{bojanowski2016enriching} vocabulary, which is reduced prior to the experiments to reduce memory usage of the model. In particular, we use the 25,000 most common words (out of about 89,000) in our dataset, which allows us to still cover 98,81\% of the occurring vocabulary. All other words get the representation $<$\textit{unk}$>$. The decoder tries to reconstruct the correct index of each word in the text from the embedding and the loss is computed using the cosine similarity between input and output feature vector.
\subsection{Classifier}
\label{sec:classifier} 
The result of the autoencoder training process is an encoder network that is able to produce highly expressive embeddings that compress visual as well as textual information to the key components, which are necessary to describe the content of the input. Based on these features we train a classifier network with our labeled (training) samples. This part of the network comprises three fully-connected layers (FC) of size (512, 512, 3), where the last one predicts one of the three different levels of \textit{Abstractness}. The entire network architecture is displayed in Figure \ref{fig:encoder_classifier}. 
\vspace{-0.7cm}

%% file: 5_experiments.tex
\section{Experimental Results}
\label{sec:experiments}
This section is separated into two parts. First, we present some example results for the CNN-based autoencoder before we present experimental results for the classification of the relative \textit{Abstractness level} of image and text. 
\vspace{-0.2cm}

\subsection{Autoencoder Training}
\begin{figure}[htbp]
	\centering
	\includegraphics[width=0.65\textwidth]{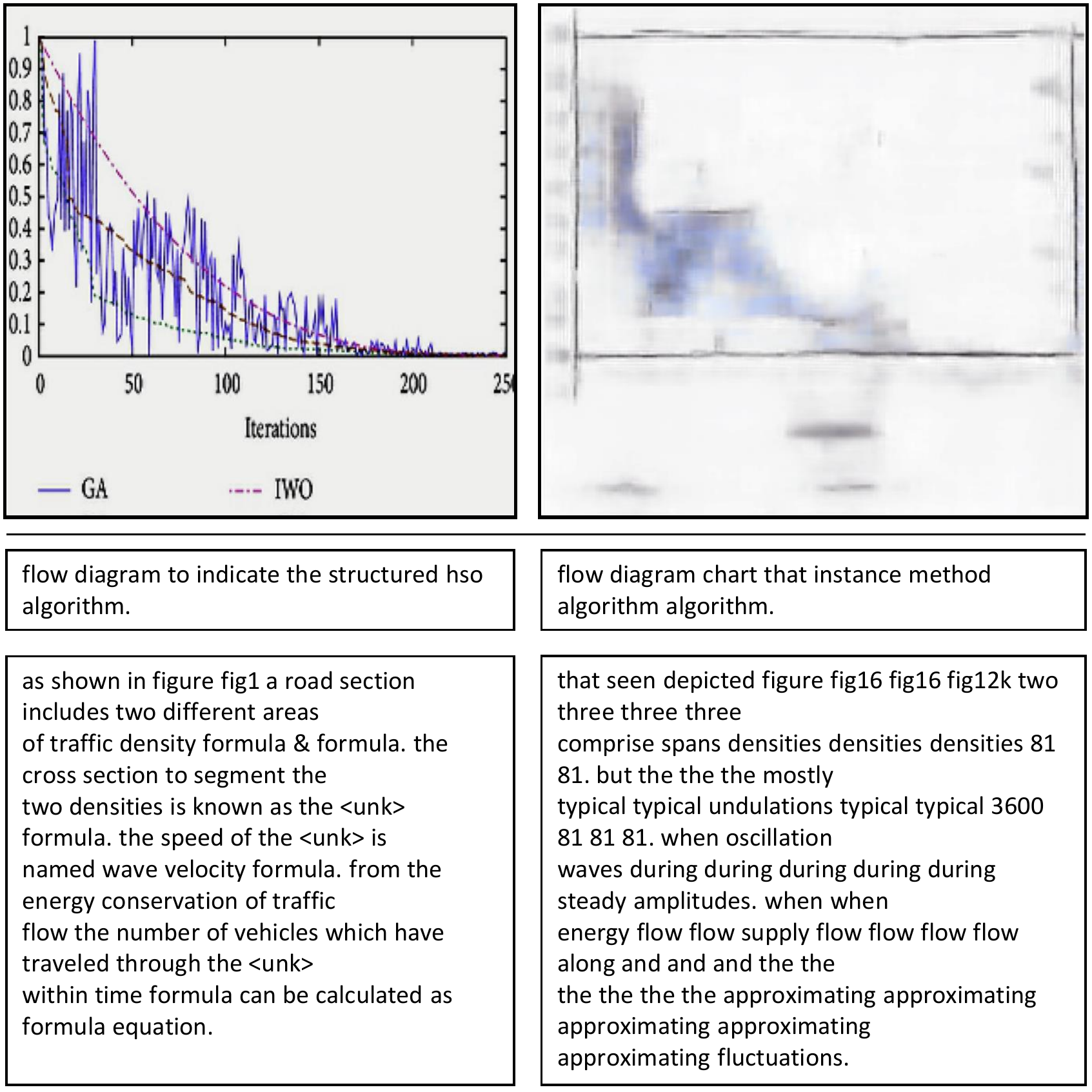}
	\caption{Three example results of the autoencoder network with the originals on the left and the reproduced samples on the right.}
	\label{fig:ae_examples}
    \vspace{-0.3cm}
\end{figure}

The autoencoder network was trained for 360,000 iterations at a batch size of 15, which corresponds to about 19 epochs. The distribution of training samples is shown in Table~\ref{fig:ae_examples} and an example output of the autoencoder is depicted in Figure \ref{fig:ae_examples}.

The example output of the autoencoder shows that the network is indeed able to coarsely reproduce the essential information of the visualizations, for instance, diagram borders, fine details on the axis, and the legend of the diagram. Also, the decoded text elements resemble the original text in length and to some extent even the semantic context.

\subsection{Classification of the Relative Abstractness Level}
\label{sec:gen_classification}
\begin{figure}[htbp]
	\centering
	\includegraphics[width=0.9\textwidth]{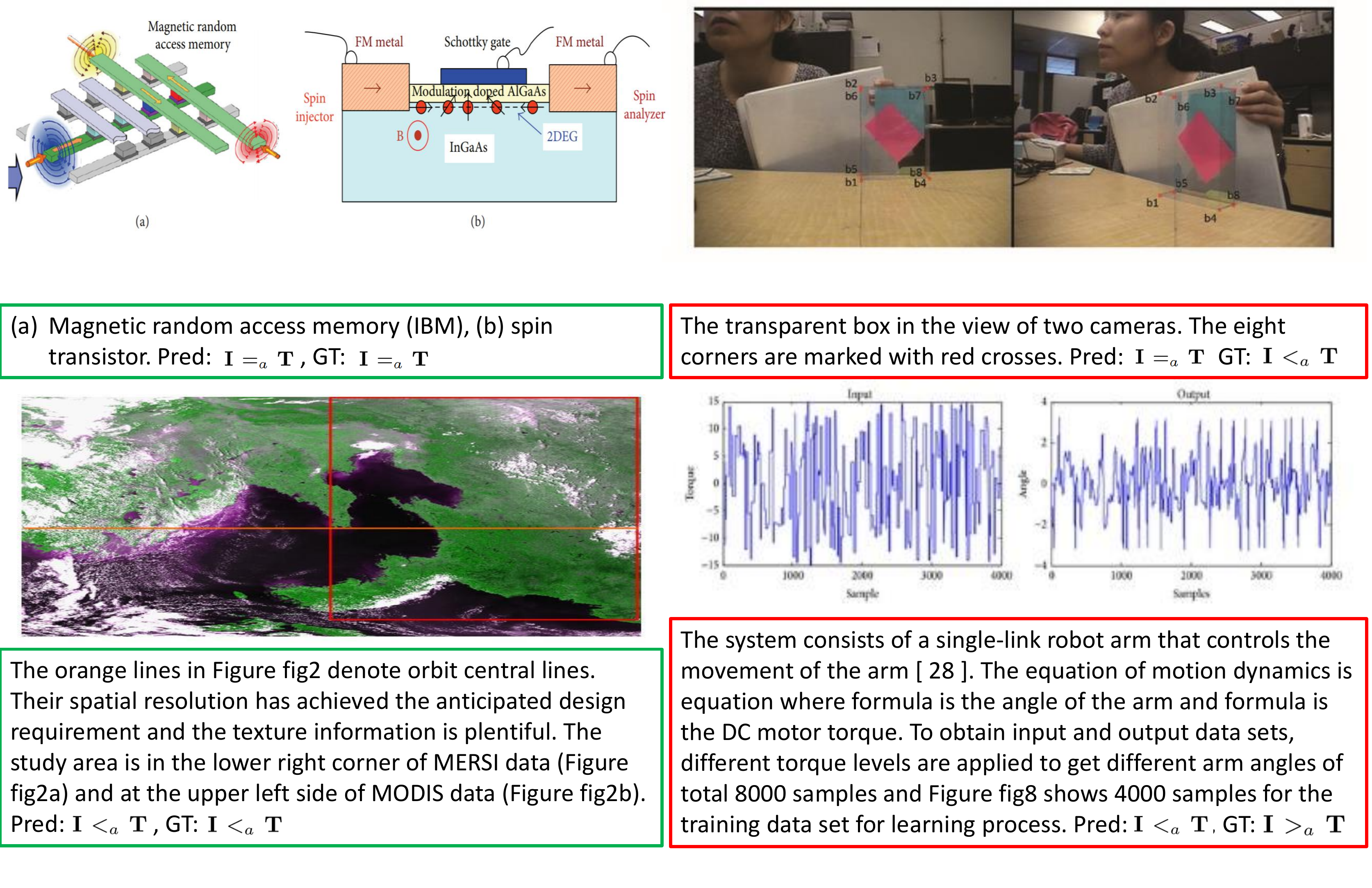}
	\caption{Examples of correctly as well as misclassified examples from our test set, along with predicted and ground-truth labels.}
	\label{fig:results_examples}
\end{figure}

As described in section \ref{sec:data}, we gathered a set of about 3,000 image-text pairs that we subsequently separated into a training and test set, where the latter one consists of 100 random samples for each of the three classes. We have evaluated three different versions of the autoencoder and classifier networks. 
\begin{enumerate}
    \item \textit{$CL_{scratch}$}: Train the classifier network as well as the encoder network from scratch, making it an end-to-end approach.
    \item \textit{$CL_{freeze}$}: Train the classifier network, but freeze the weights of the pre-trained encoder network.
    \item \textit{$CL_{transfer}$}: Train the classifier network and finetune the pre-trained encoder network at the same time.
\end{enumerate}
\begin{table}[htbp]
	\centering
	\begin{tabular}{| l | c | c | c | c |}
		\hline
		Class & Image $<_a$ Text & Image $>_a$ Text & Image $=_a$ Text & Sum\\ 
		\hline
		I $<_a$ T & \textbf{90} & 7           & 3            & 100\\ 
		\hline
		I $>_a$ T & 14          & \textbf{68} & 18           & 100\\ 
		\hline
		I $=_a$ T  & 10         & 7           & \textbf{83}  & 100\\
		\hline \hline
		Precision & $78.95\%$ & $82.93\%$ & $79.81\%$ & -\\ \hline 
		Recall & $90.00\%$ & $68.00\%$ & $83.00\%$ & -\\ \hline
	\end{tabular}
	\centering	
	\vspace{0.1cm}
    \caption{Confusion matrix for the classifier of the relative abstraction level.}
	\label{tab:confusion_matrix_classifier}
\vspace{-0.3cm}
\end{table}
\begin{table}[htbp]
	\begin{tabular}{| l | c | c | c |}
		\hline
		Classifier & $CL_{transfer}$ & $CL_{freeze}$ & $CL_{scratch}$\\ 
		\hline
		Accuracy   & \textbf{$80.33\%$} & $77.33$ & $77.00$ \\ 
		\hline
	\end{tabular}
	\centering
	\vspace{0.1cm}
	\caption{Comparison of the three different classification approaches.}
	\label{tab:classifier_comparison}
\vspace{-0.3cm}
\end{table}
Every approach was trained for about 70,000 iterations. The results are reported in Tables \ref{tab:confusion_matrix_classifier} and \ref{tab:classifier_comparison}. The former shows that the classifier was able to predict the three classes successfully with a recall of 90\% ($I<_{a}T$), 68\% ($I>_{a}T$) and 83\% ($I=_{a}T$). These results reflect the distribution of available labeled training data (cf. Table~\ref{tab:labeledset}), which implies that these results can be improved by acquiring more annotated samples. Table~\ref{tab:classifier_comparison} shows that a pre-trained autoencoder network outperforms a training from scratch, but is in turn outperformed by the transfer learning approach which finetunes and adapts the encoding process to the new task. This proves that a multimodal embedding is able to encode an image-text pair in a way that the \textit{Abstractness} metric can be successfully predicted, while our autoencoder approach is able to compensate for the relatively low number of labeled training samples. Example predictions of our system are displayed in Figure~\ref{fig:results_examples} (left hand side: correct, right hand side: misclassified): In the top-left image-text pair, both the text and the schematic illustrations are abstract representations, in particular for "(a)", while in the image bottom-left the image is a 	concretization of the text (both predicted correctly); the image in the top-right examples depicts more relevant details than the text, whereas the line chart bottom-right provides less detailed information about the experimental context than the text. As Table~\ref{tab:confusion_matrix_classifier} shows, predicting $I<_{a}T$ was the easiest for the system, presumably because of the amount of natural images in this class. But, if these natural images were overlaid with additional information (top right) the system struggled to find the correct assignment.

%% file: 6_conclusion.tex
\section{Conclusions}
\label{sec:conclusion}

In this paper, we have introduced a novel metric that describes the relative \textit{Abstractness Level} between an image and associated text. We have motivated and derived the metric based on previous work on taxonomies for image-text classes in communication sciences and applied linguistics. Until now, the large variety of image-text relations had been investigated in a differentiated way mainly in these fields. 
We have set these taxonomies in relation to recent work in the field of multimedia retrieval, which has modeled image-text relations in a more general manner through the metrics \textit{cross-modal mutual information} and \textit{semantic correlation}.
In this respect, our proposed metric is a contribution to model the variety of possible semantic (cross-modal) image-text relations in a systematic manner from an information retrieval perspective. Moreover, we have proposed a deep learning architecture to automatically predict the relative, cross-modal abstractness level of image and text. The required amount of labeled training data is minimized by the incorporation of an autoencoder network. We have evaluated three different ways of training the deep network architecture. It turned out that training the classifier network and finetune the pre-trained encoder network at the same time achieved the best results with an accuracy of 80~\%. In this way, an indexing method has been developed that can serve as the basis for multimodal search and retrieval, for instance, in order to search for educational and scientific content.  

In the future, we plan to apply this indexing method to different scenarios in multimodal information retrieval, such as search as learning with multimedia data, e-learning, and recommender systems. For this purpose we intend to build an exploration and browsing interface 
based on the metrics \textit{CMI}, \textit{SC} and \textit{ABS}. Finally, we will evaluate the usefulness of other metrics to model cross-modal relations in a systematic way.

%% file: main.bbl
\begin{thebibliography}{10}
\providecommand{\url}[1]{\texttt{#1}}
\providecommand{\urlprefix}{URL }
\providecommand{\doi}[1]{https://doi.org/#1}

\bibitem{afouras2018deep}
Afouras, T., Chung, J.S., Senior, A., Vinyals, O., Zisserman, A.: Deep
  audio-visual speech recognition. arXiv preprint arXiv:1809.02108  (2018)

\bibitem{ba2016layer}
Ba, J.L., Kiros, J.R., Hinton, G.E.: Layer normalization. arXiv preprint
  arXiv:1607.06450  (2016)

\bibitem{balaneshin2018deep}
Balaneshin-kordan, S., Kotov, A.: Deep neural architecture for multi-modal
  retrieval based on joint embedding space for text and images. In: Proceedings
  of the Eleventh ACM International Conference on Web Search and Data Mining.
  pp. 28--36. ACM (2018)

\bibitem{baltruvsaitis2018multimodal}
Baltru{\v{s}}aitis, T., Ahuja, C., Morency, L.P.: Multimodal machine learning:
  A survey and taxonomy. IEEE Transactions on Pattern Analysis and Machine
  Intelligence  (2018)

\bibitem{barthes1977image}
Barthes, R.: Image-music-text, ed. and trans. S. Heath, London: Fontana
  \textbf{332} (1977)

\bibitem{bateman2014text}
Bateman, J.: Text and image: A critical introduction to the visual/verbal
  divide. Routledge (2014)

\bibitem{bojanowski2016enriching}
Bojanowski, P., Grave, E., Joulin, A., Mikolov, T.: Enriching word vectors with
  subword information. arXiv preprint arXiv:1607.04606  (2016)

\bibitem{bucak2014multiple}
Bucak, S.S., Jin, R., Jain, A.K.: Multiple kernel learning for visual object
  recognition: A review. IEEE Transactions on Pattern Analysis and Machine
  Intelligence  \textbf{36}(7),  1354--1369 (2014)

\bibitem{carvalho2018cross}
Carvalho, M., Cad{\`e}ne, R., Picard, D., Soulier, L., Thome, N., Cord, M.:
  Cross-modal retrieval in the cooking context: Learning semantic text-image
  embeddings. arXiv preprint arXiv:1804.11146  (2018)

\bibitem{cho2014learning}
Cho, K., van Merrienboer, B., Gulcehre, C., Bahdanau, D., Bougares, F.,
  Schwenk, H., Bengio, Y.: {Learning Phrase Representations using RNN
  Encoder-Decoder for Statistical Machine Translation}. Association for
  Computational Linguistics (2014)

\bibitem{Fan2017}
Fan, M., Wang, W., Dong, P., Han, L., Wang, R., Li, G.: Cross-media retrieval
  by learning rich semantic embeddings of multimedia. ACM Multimedia Conference
   (2017)

\bibitem{gonen2011multiple}
G{\"o}nen, M., Alpayd{\i}n, E.: Multiple kernel learning algorithms. Journal of
  machine learning research  \textbf{12}(Jul),  2211--2268 (2011)

\bibitem{halliday2013halliday}
Halliday, M.A.K., Matthiessen, C.M.: Halliday's introduction to functional
  grammar. Routledge (2013)

\bibitem{henning2017estimating}
Henning, C.A., Ewerth, R.: Estimating the information gap between textual and
  visual representations. ACM International Conference on Multimedia Retrieval
  (2017)

\bibitem{jaques2015multi}
Jaques, N., Taylor, S., Sano, A., Picard, R.: Multi-task, multi-kernel learning
  for estimating individual wellbeing. In: Proc. NIPS Workshop on Multimodal
  Machine Learning, Montreal, Quebec. vol.~898 (2015)

\bibitem{jia2015guiding}
Jia, X., Gavves, E., Fernando, B., Tuytelaars, T.: Guiding the long-short term
  memory model for image caption generation. In: Proceedings of the IEEE
  International Conference on Computer Vision. pp. 2407--2415 (2015)

\bibitem{jin2016video}
Jin, Q., Liang, J.: Video description generation using audio and visual cues.
  In: Proceedings of the 2016 ACM on International Conference on Multimedia
  Retrieval. pp. 239--242. ACM (2016)

\bibitem{kang2015cross}
Kang, C., Liao, S., He, Y., Wang, J., Niu, W., Xiang, S., Pan, C.: Cross-modal
  similarity learning: A low rank bilinear formulation. In: ACM Conference on
  Information and Knowledge Management. ACM (2015)

\bibitem{Liang2016}
Liang, J., Li, Z., Cao, D., He, R., Wang, J.: Self-paced cross-modal subspace
  matching (2016)

\bibitem{liu2014multiple}
Liu, F., Zhou, L., Shen, C., Yin, J.: Multiple kernel learning in the primal
  for multimodal alzheimer's disease classification. IEEE J. Biomedical and
  Health Informatics  \textbf{18}(3),  984--990 (2014)

\bibitem{marsh2003taxonomy}
Marsh, E.E., Domas~White, M.: A taxonomy of relationships between images and
  text. Journal of Documentation  \textbf{59}(6),  647--672 (2003)

\bibitem{Martinec2005}
Martinec, R., Salway, A.: {A system for image-text relations in new (and old)
  media}. Visual Communication  \textbf{4} (2005)

\bibitem{meutzner2017improving}
Meutzner, H., Ma, N., Nickel, R., Schymura, C., Kolossa, D.: Improving
  audio-visual speech recognition using deep neural networks with dynamic
  stream reliability estimates. In: Acoustics, Speech and Signal Processing
  (ICASSP), 2017 IEEE International Conference on. pp. 5320--5324. IEEE (2017)

\bibitem{neverova2016moddrop}
Neverova, N., Wolf, C., Taylor, G., Nebout, F.: Moddrop: adaptive multi-modal
  gesture recognition. IEEE Transactions on Pattern Analysis and Machine
  Intelligence  \textbf{38}(8),  1692--1706 (2016)

\bibitem{poria2015deep}
Poria, S., Cambria, E., Gelbukh, A.: Deep convolutional neural network textual
  features and multiple kernel learning for utterance-level multimodal
  sentiment analysis. In: Proceedings of the 2015 conference on empirical
  methods in natural language processing. pp. 2539--2544 (2015)

\bibitem{rajagopalan2016extending}
Rajagopalan, S.S., Morency, L.P., Baltrusaitis, T., Goecke, R.: Extending long
  short-term memory for multi-view structured learning. In: European Conference
  on Computer Vision. pp. 338--353. Springer (2016)

\bibitem{Ramanishka2016}
Ramanishka, V., Das, A., Park, D.H., Venugopalan, S., Hendricks, L.A.,
  Rohrbach, M., Saenko, K.: {Multimodal Video Description} (2016)

\bibitem{Shutova2016}
Shutova, E., Kelia, D., Maillard, J.: {Black Holes and White Rabbits : Metaphor
  Identification with Visual Features}. Naacl  (2016)

\bibitem{sohmen2018figures}
Sohmen, L., Charbonnier, J., Bl{\"u}mel, I., Wartena, C., Heller, L.: Figures
  in scientific open access publications. In: International Conference on
  Theory and Practice of Digital Libraries. pp. 220--226. Springer (2018)

\bibitem{szegedy2017inception}
Szegedy, C., Ioffe, S., Vanhoucke, V., Alemi, A.A.: Inception-v4,
  inception-resnet and the impact of residual connections on learning. In:
  AAAI. vol.~4, p.~12 (2017)

\bibitem{szegedy2015going}
Szegedy, C., Liu, W., Jia, Y., Sermanet, P., Reed, S., Anguelov, D., Erhan, D.,
  Vanhoucke, V., Rabinovich, A.: Going deeper with convolutions (2015)

\bibitem{unsworth2007image}
Unsworth, L.: Image/text relations and intersemiosis: Towards multimodal text
  description for multiliteracies education. In: Proceedings of the 33rd
  International Systemic Functional Congress. pp. 1165--1205 (2007)

\bibitem{vinyals2015show}
Vinyals, O., Toshev, A., Bengio, S., Erhan, D.: Show and tell: A neural image
  caption generator. pp. 3156--3164 (2015)

\bibitem{yan2016supervised}
Yan, T.K., Xu, X.S., Guo, S., Huang, Z., Wang, X.L.: Supervised robust discrete
  multimodal hashing for cross-media retrieval. In: ACM Conference on
  Information and Knowledge Management (2016)

\bibitem{yang2016hierarchical}
Yang, Z., Yang, D., Dyer, C., He, X., Smola, A., Hovy, E.: Hierarchical
  attention networks for document classification. In: Proceedings of the 2016
  Conference of the North American Chapter of the Association for Computational
  Linguistics: Human Language Technologies. pp. 1480--1489 (2016)

\bibitem{yeh2012novel}
Yeh, Y.R., Lin, T.C., Chung, Y.Y., Wang, Y.C.F.: A novel multiple kernel
  learning framework for heterogeneous feature fusion and variable selection.
  IEEE Transactions on multimedia  \textbf{14}(3),  563--574 (2012)

\end{thebibliography}
